\documentclass{article}

\usepackage{microtype}
\usepackage{graphicx}
\usepackage{subfigure}
\usepackage{booktabs} 

\usepackage{amsmath,amsfonts,bm}

\def\eqref#1{equation~\ref{#1}}

\def\1{\bm{1}}

\def\vtheta{{\bm{\theta}}}
\def\va{{\bm{a}}}

\def\ve{{\bm{e}}}
\def\vf{{\bm{f}}}

\def\vl{{\bm{l}}}

\def\vp{{\bm{p}}}
\def\vq{{\bm{q}}}

\def\vs{{\bm{s}}}

\def\vu{{\bm{u}}}
\def\vv{{\bm{v}}}

\def\vx{{\bm{x}}}
\def\vy{{\bm{y}}}

\def\mI{{\bm{I}}}
\def\mJ{{\bm{J}}}

\def\mM{{\bm{M}}}

\def\mR{{\bm{R}}}

\def\mT{{\bm{T}}}

\DeclareMathAlphabet{\mathsfit}{\encodingdefault}{\sfdefault}{m}{sl}
\SetMathAlphabet{\mathsfit}{bold}{\encodingdefault}{\sfdefault}{bx}{n}

\usepackage[numbers]{natbib}
\usepackage{multicol}
\usepackage[bookmarks=true]{hyperref}
\usepackage{amsmath}
\usepackage{amssymb}
\usepackage{graphicx}
\usepackage{xcolor}
\usepackage{lscape}
\usepackage{booktabs}
\usepackage{mathtools}
\usepackage{booktabs}
\def\tran{^\intercal}

\usepackage{hyperref}

\usepackage[accepted]{icml2020}

\icmltitlerunning{A Differentiable
Newton Euler
Algorithm
for
Multi-body
Model Learning}

\begin{document}

\twocolumn[
\icmltitle{A Differentiable
Newton Euler
Algorithm
for
Multi-body
Model Learning}



\icmlsetsymbol{equal}{*}

\begin{icmlauthorlist}
\icmlauthor{Michael Lutter}{equal,tuda}
\icmlauthor{Johannes Silberbauer}{equal,tuda}
\icmlauthor{Joe Watson}{tuda}
\icmlauthor{Jan Peters}{tuda}
\end{icmlauthorlist}

\icmlaffiliation{tuda}{Computer Science Department, Technical University of Darmstadt, Darmstadt, Germany}

\icmlcorrespondingauthor{Michael Lutter}{michael@robot-learning.de}

\icmlkeywords{Machine Learning, ICML}

\vskip 0.3in
]



\printAffiliationsAndNotice{\icmlEqualContribution} 

\vspace{-0.6em}
\section{Introduction} \vspace{-0.6em}
\noindent
The identification of dynamical systems from data is a powerful tool in Robotics \cite{aastrom1971system}.
Learnt analytic models may be used for control synthesis and can be utilized for gravitational and inertial compensation \cite{spong2020robot}. 
Moreover, when used as simulators, they can be used to reduce the sample complexity of data-driven control methods such as Reinforcement Learning \cite{deisenroth2011pilco, chua2018deep}.
For these control applications, where out-of-distribution prediction is typically required, the ability to generalize beyond the acquired data is critical.
Any modelling error may be exploited by a controller, and such exploitation may result in catastrophic system failure.
To ensure sufficient out-of-sample generalization, the model's hypothesis space is an important consideration.
Ideally, this space should be defined such that only plausible trajectories, that are physically consistent and 
have bounded energy, are generated. 

\medskip
\noindent Standard black-box models such as deep networks or Gaussian processes, which are a common choice for model learning \cite{nguyen2011survey, deisenroth2011pilco, chua2018deep}, have a broad hypothesis space and so can simulate dynamics with unbounded energy.
To overcome this shortcoming, `grey-box' models that combine deep networks with physical insights have been recently proposed, e.g., incorporating Lagrangian \cite{lutter2018deep, Lutter2019Energy, gupta2019general} and Hamiltonian Mechanics \cite{greydanus2019hamiltonian} for energy-conserving models.
These empirically show better generalization compared to black-box models, but the local representations of deep networks cannot guarantee out-of-sample generalization.
Only white-box models \cite{atkeson1986estimation, ting2006bayesian, traversaro2016identification, wensing2017linear, sutanto2020encoding}, which infer the physical parameters of the system given the analytic equations of motion, can guarantee out-of sample generalization as these models are valid globally. 
While these combinations of physics with data-driven learning can obtain more robust representations, the usage of physics priors commonly reduces model accuracy compared to black-box methods as most physics priors cannot capture reality to a sufficient fidelity.
The priors of Newtonian-, Lagrangian- and Hamiltonian mechanics typically cannot describe the complex nonlinear phenomena of friction, hysteresis and contact. Therefore, the need of a compromise between physics-inspired models and black-box functions approximation is clear. 
 
\medskip 
\noindent In this work, we examine a spectrum of hybrid models for the domain of multibody robot dynamics.
We motivate a computation graph architecture that embodies the Newton Euler equations, emphasising the utility of the Lie Algebra form in translating the dynamical geometry into an efficient computational structure for learning \cite{handa2016gvnn}.
We describe the used actuator models (Section \ref{sec:actuators}) and the virtual parameters (Appendix).
In the experiments, we evaluate 26 Newton-Euler based system identification approaches and benchmark these models on the simulated and physical Furuta Pendulum and Cartpole.
The comparison shows that the kinematic parameters, required by previous Newton-Euler methods \cite{atkeson1986estimation, sutanto2020encoding, ledezma2017first}, can be accurately inferred from data.
Furthermore, we highlight that models with guaranteed bounded energy of the uncontrolled system generate non-divergent trajectories, while more general models have no such guarantee. Therefore, their performance strongly depends on the data distribution.
The main contributions of this work are the introduction of a white-box model that jointly learns dynamic and kinematics parameters and can be combined with black-box components.
We then provide an extensive empirical evaluation on challenging systems and different datasets that elucidates the comparative performance of our grey-box architecture with comparable white- and black-box models.

\section{Newton Euler Equations} \vspace{-0.6em}
\noindent Consider a set of coupled rigid bodies, such as a robotic manipulator.
While the global dynamics of such a system is complex, the behaviour of each individual body and its effect on its neighbours are easily understood.
This is the key insight of algorithms such as Recursive Newton Euler Algorithm (RNEA) and Articulated Body (ABA) \cite{Featherstone2007rigid}, which efficiently solve for the global dynamics by successively propagating the local solutions forward and backwards.
While alternative methods for modelling, such as Lagrangian and Hamiltonian Mechanics, may be appreciated for their mathematical elegance and convenience; the Newton Euler approach is based on global and interpretable physical quantities enabling robust out-of-sample generalization.
Moreover, the efficient algorithmic form is applicable for both forward and inverse dynamics
We describe the Newton Euler method in its Lie Algebra form \cite{kim2012lie}, which is not only compact, but easy to represent as a differentiable computation graph \cite{handa2016gvnn}.

\medskip
\noindent In this paper we focus on kinematic trees composed on $n$ components.
Given a base reference frame, the pose of each body can be described via successive application of an affine transform $\mT$ containing the 3x3 rotation matrix $\mR$ and translation $\vp{\in}\mathbb{R}^3$.
To describe the dynamics of each body, we use a 'generalised' velocity $\bar{\vv} {=}[\vv, \bm{\omega}]\tran$, which is composed of linear ($\vv$) and rotational ($\bm{\omega}$) components. The generalised velocity $\bar{\vv}_i$ expresses the link velocity in the inertial base frame but expressed in the $i$th link coordinate frame. The generalised notation is extended to acceleration $\bar{\va}$, force $\bar{\vf}{=}[\vf, \bm{\tau}]\tran$ (with moment $\bm{\tau}$), momentum $\bar{\vl}$,  and inertia 
\begin{align}
    \bar{\mM} &= 
    \begin{bmatrix}
        \mJ       & m[\vp_m] \\
         m[\vp_m]\tran & m\mI
    \end{bmatrix}
\end{align}
with link mass $m$, link inertia $\mJ$ and link center of mass (CoM) position $\vp_m$ relative to the body frame, and $[\vx]$ mapping the vector $\vx$ to a skew-symmetric matrix such that $\vx{\times}\vy{=}[\vx]\vy$.
The benefit of the Lie Algebra formulation is apparent when considering the kinematics, dynamics and differential equations.
Transforming the kinematic terms (e.g. $\bar{\vv}, \bar{\va}$) of the $i$th body in the $j$th frame is a compact linear operations through use of the adjoint transform of $\mT$ ($\text{Ad}_{\mT}$), i.e.,
\begin{align}
    \bar{\vv}_{j} &=
    \text{Ad}_{\mT_{j,i}}\bar{\vv}_i,
    \hspace{1.5em}
    \bar{\va}_{j}=
    \text{Ad}_{\mT_{j,i}}\bar{\va}_i.
\end{align}
In the Lie algebra formulation, the kinematic terms can be described in the SE(3) manifold due to the properties of the affine transform $\mT$. 
Dynamic terms (e.g. $\bar{\vl}, \bar{\vf}$) can be shown to act in the \textit{dual space} for SE(3), DSE(3).
Hence, these terms can be transformed between frames using the coadjoint operator $\text{Ad}^*_{\mT_{j,i}}{=}\text{Ad}\tran_{\mT_{j,i}}$, i.e.,
\begin{align}
    \bar{\vl}_{j}= \text{Ad}^*_{\mT_{j,i}}\bar{\vl}_i,
    \hspace{1.5em}
    \bar{\vf}_{j}= \text{Ad}^*_{\mT_{j,i}}\bar{\vf}_i.
\end{align}
Differential equations in the inertial base frame can also be expressed compactly, i.e., the Newton-Euler equation is described by
\begin{align*}
    \bar{\vf} = \frac{d}{dt}\bar{\vl} &=
    \bar{\mM}\bar{\va} - \text{ad}^*_{\bar{\vv}}\bar{\mM}\bar{\vv},
    \hspace{5pt}  \text{where} \hspace{5pt} 
    \text{ad}^*_{\bar{\vv}} = 
    \begin{bmatrix}
            [\bm{\omega}]   & \mathbf{0} \\
         [\vv] & [\bm{\omega}]
    \end{bmatrix}.
\end{align*}
These transformations and differential equations can be embedded in articulated body algorithm (ABA) and the Recursive Newton Euler Algorithm (RNEA). These different formulation yield more compact and intuitive description compared to the original formulation \cite{Featherstone2007rigid}, e.g., the ABA algorithm in Lie algebra is described in Algorithm \ref{alg:articulated-rigid-body}. 

\section{Actuator Models} \label{sec:actuators} \vspace{-0.6em}
\noindent The models of the previous sections mostly focused on simulating rigid body dynamics. This representation does commonly not capture reality with  sufficient fidelity for mechanical systems with actuation. Actuators exerting the control signal are affected by non-linear transformation of the set-points, hysteresis and friction. To be able to learn such models with non-ideal actuators, we augment the rigid-body dynamics model with an actuator model. This actuation model can either be white-box model relying on existing friction models or black-box models. We define six different joint independent actuator models, building from $\tau =  \tau_d$,
\begin{align*}
    \text{Viscous:}& &\tau &= \tau_d {-} \mu_v \dot{q}, \\
    \text{Stribeck:}& &\tau &= \tau_d {-} \text{sign}(\dot{q}) \left(f_s {+} f_d \exp\left({-}\nu_s \dot{q}^2 \right)\right) {-} \mu_v \dot{q},\\
    \text{NN Friction:}& &\tau &= \tau_d {-} \text{sign}(\dot{q}) \: \| f_{\text{NN}}(q,\dot{q}; \psi_F) \|_1 ,\\
    \text{NN Residual:}& &\tau &= \tau_d {-} f_{\text{NN}}(q,\dot{q}; \psi_R), \\
    \text{FF-NN:}& &\tau &= f_{\text{NN}}(\tau_d, q,\dot{q}; \psi_M),
\end{align*}
where $\tau_d$ is the desired torque. The Viscous, Stribeck and MLP friction actuator models are guaranteed to learn a stable uncontrolled system (i.e., $\tau_d{\coloneqq}0$) as all possible actuator parameters can only dissipate energy, i.e., $\dot{E} = \bm{\tau}^T \dot{\vq} \leq 0 \hspace{4pt} \forall \hspace{4pt} \theta_F$.
The Residual and MLP actuator models are capable of generating energy and so the performance will depend on the training data.

\section{Experimental Setup} \vspace{-0.6em}
\noindent In the following the experimental setup containing the used systems, datasets and models is described.

\subsection{Systems} \vspace{-0.6em}
\noindent For the comparison we choose the Quanser Cartpole and the Quanser Furuta Pendulum as these systems are under-actuated and the physical systems have different peculiarities, which make system identification challenging. System identification for under-actuated systems is harder compared to fully-actuated systems as all data points lie on a low dimensional manifold and one cannot capture the whole space.

\begin{figure*}[t]
    \centering
    \includegraphics[width=\textwidth]{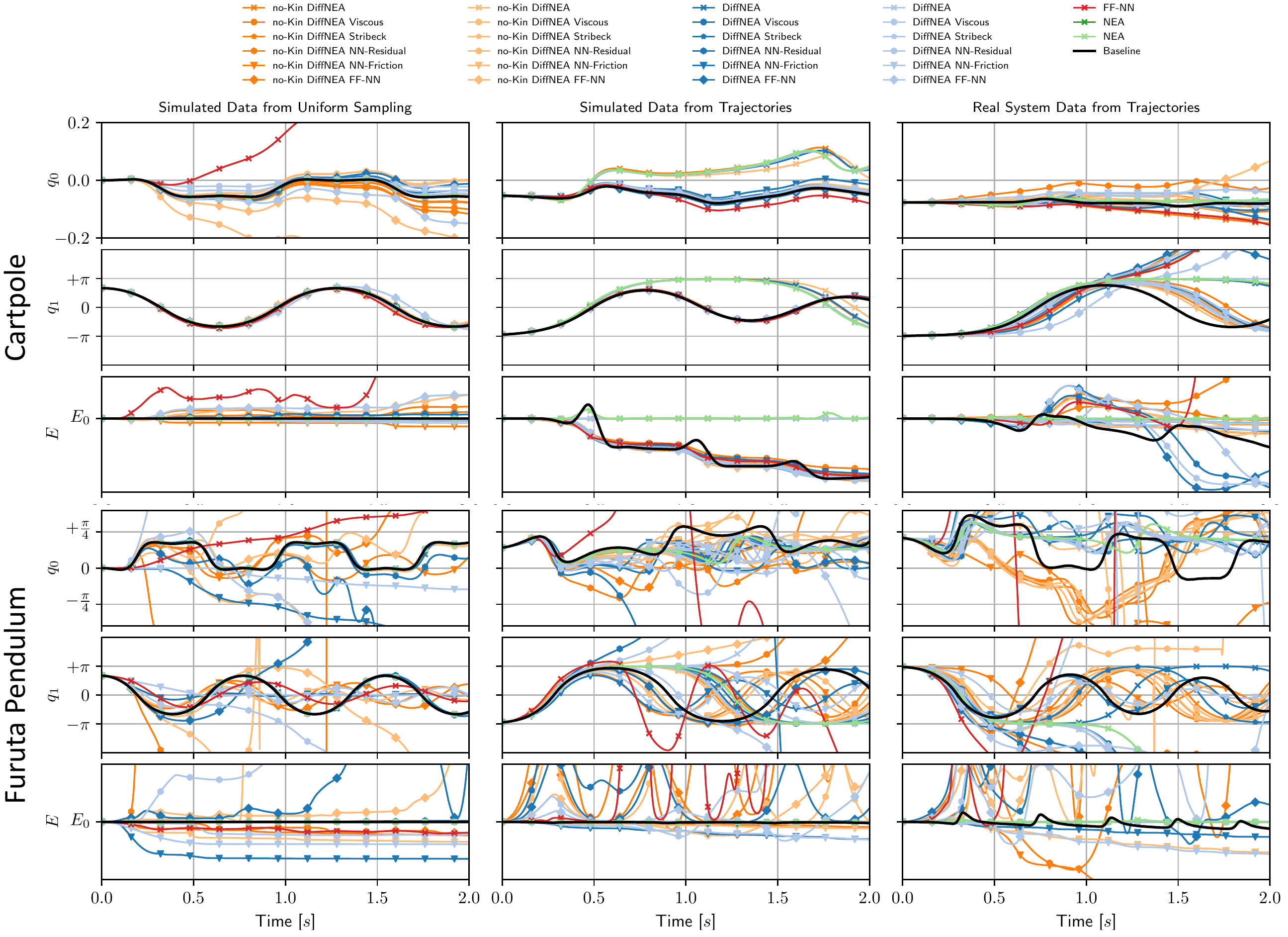}
    \caption{Qualitative model comparison of the 26 different models performing forward roll-outs on the Cartpole and Furuta Pendulum. The roll-outs start from the starting state and are computed with 250Hz sampling frequency and integrated with RK4. The models are trained on three different datasets ranging from uniformly sampled and ideal observations (i.e., Simulated Data from Uniform Sampling) to trajectory data of noisy observation from the physical system.}
    \label{fig:rollout}
\end{figure*}

\subsection{Datasets} \vspace{-0.6em}
\noindent To evaluate the impact of data quality, we are evaluating the performance on three different datasets with different levels of complexity. The simulated data from uniform sampling dataset is generated by sampling joint positions, velocities and torques from a uniform distribution spanning the complete state domain and computing the acceleration with the true analytic forward dynamics. The simulated data from trajectories dataset is generated by simulating the ideal system with viscous friction and small state and action noise. The real system data from trajectories dataset is generated on the physical system by applying an energy controller that repeatedly swings-up the pendulum and lets it fall down without actuation. 

\subsection{Models} \vspace{-0.6em}
\noindent For the evaluation we compare three different instantiations of the  previously described white-box model family with the 6 different actuation models, if applicable.

\textbf{No-Kin Differential Newton Euler Alg.:}
The no-Kin DiffNEA model assumes knowledge of the kinematic tree but no knowledge of the kinematics $\vtheta_K$ or dynamics parameters $\vtheta_I$. These link parameters are learned from data containing only generalized coordinates using ADAM to minimize the squared loss of the forward dynamics. 

\textbf{Differential Newton Euler Alg.:}
The DiffNEA model assumes knowledge of the kinematic chain and the kinematics parameters $\vtheta_K$ and only learns $\vtheta_I$. These parameters are learnt by minimizing the squared loss of the forward dynamics using ADAM. This approach was introduced by \cite{sutanto2020encoding}.

\textbf{Newton Euler Alg.:}
The NEA model assumes knowledge of the kinematic chain and the kinematics parameters $\vtheta_K$ and only learns $\vtheta_I$. These parameters are learnt by linear regression. This model learning approach was initially introduced by \cite{atkeson1986estimation}. Due to the linear regression this model cannot be augmented with the different actuation models.

\textbf{Feed-Forward Neural Network (FF-NN)}
As black-box model learning baseline, we are using a feed-forward neural network trained via ADAM. This network is a continuous time model and predicts the joint acceleration.
 \vspace{-0.3em}
\subsection{Model Initialization}  \vspace{-0.6em}
For the white box models we are differentiating between two initialization strategies, without prior and with prior. Without prior means that the link parameters are initialized randomly. With prior means that the parameters are initialized with the known values given by the manufacturer. This differentiation enables us to evaluate the impact of good initialization for white box models. 

\section{Experimental Results}\vspace{-0.6em}
\noindent The qualitative experimental results are shown in Figure \ref{fig:rollout}. The overall performance depends heavily on the system as well as the dataset.
The numerically sensitive conditioning of the Furuta Pendulum causes all models to be worse on all datasets and model classes.
Conversely, the magnitude of the physical parameters of the Cartpole make the identification and long-term prediction is simpler.
Regarding the datasets, the overall forward prediction performance, with some outliers, decreases with dataset complexity.
Regarding the different models, no clear best system identification representation and identification approach from those studied is apparent.
One interesting result is that the white-box model approach with \emph{unknown} kinematics (no-Kin DiffNEA) performs comparable to the white-box model with \emph{known} kinematics (DiffNEA, NEA), demonstrating the kinematics and dynamics can be learned jointly.
Furthermore, one can observe that the three different model classes, energy-conserving, energy-bounded and energy-unbounded models achieve very different long-term forward prediction.

\subsection{Energy Conserving Models} \vspace{-0.6em}
\noindent The energy-conserving models, i.e. no-Kin DiffNEA, DiffNEA, NEA and no-actuator model, can only represent energy conserving dynamics.
When this prior is correct, as in the simulated uniform dataset, these models perform well.
If the prior is not correct, e.g., the large viscous friction of the \emph{simulated} Cartpole, the forward prediction degrades significantly even for simulated data.
The identified parameters are physically plausible but unreasonable. While the mass of the cart is about $0.5$kg, 
this model class sets the mass to $1$kg 
for the simulation and $2$kg 
for the physical system. The mass of the about $120$g  
pendulum is set to  $340$g  
for the simulation and $3000$g  
for the physical system.

\vspace{-0.3em}
\subsection{Energy Bounded Models}\vspace{-0.6em}
\noindent The energy-bounded models, i.e., NEA plus Viscous, Stribeck and NN-Friction actuator, guarantee that the energy does not increase without actuation and hence, guarantee to learn a global stable system. This model class contains the best performing models of this benchmark.
The learnt models always yield non-diverging trajectories and can model systems with and without friction.
The NN-Friction actuator achieves particularly good performance by exploiting its black-box flexibility within the inductive bias.
Despite its expressiveness, the model does not overfit and the obtained physical parameters are comparable to the white-box actuator models. 

\vspace{-0.3em}
\subsection{Energy Unbounded Models} \vspace{-0.6em}
\noindent The energy-unbounded models, i.e., FF-NN, NEA plus FF-NN and NN-Residual actuator, can potentially learn to increase the system energy during simulation without actuation, which is physically implausible.
The benchmark of Figure \ref{fig:rollout} shows that models of this class learn to pump energy into the system even for perfect sensor measurements (i.e., FF-NN of the Cartpole with simulated uniformly sampled data).
For the more challenging trajectory datasets, all black-box and hybrid models learn models that increase in energy during simulation without actuation.
Many of these models also generate divergent trajectories during simulation, which leave the training domain. 

\vspace{-0.3em}
\subsection{Model Initialization} \vspace{-0.6em}
\noindent For the the hybrid white-box models the different model initializations were compared.
In the simulation experiments no clear difference between models with and without prior initialization is visible. Evaluating the identified physical parameters also yields no clear improvement of the initialization with prior, e.g., even for unreasonably large physical parameters, the identified parameter with a prior was not necessarily smaller.
Therefore, we conclude that the initialization with the best known parameters does not necessarily improve model performance when using stochastic gradient descent with ADAM.  

\vspace{-0.3em}
\section{Conclusion} \vspace{-0.6em}
\noindent In this paper, we have described the classical Newton-Euler system identification approach \cite{atkeson1986estimation} using the elegant Lie algebra formulation \cite{kim2012lie} and the differential programming paradigm. We combined this formulation with white- and black-box actuation models for end-to-end learning.
In a large-scale benchmark, we compared 26 different models on three datasets and two different systems.
The two main conclusions of this benchmark are, (1) the no-Kin DiffNEA model, which learns the kinematics, performs equally well compared to the same DiffNEA with the kinematics given, and (2) models with guaranteed global stability yield the best long term forward simulations. 

\section*{Acknowledgment}
This project has received funding from the European Union’s Horizon 2020 research and innovation program under grant agreement No \#640554 (SKILLS4ROBOTS). Furthermore, this research was also supported by grants from ABB, NVIDIA and the NVIDIA DGX Station.


\bibliography{references}
\bibliographystyle{icml2020}

\appendix
\section*{Appendix}
\begin{algorithm}[h]
\caption{Forward dynamics with Articulated Rigid Body for a kinematic tree in terms of Lie algebra.}
\label{alg:articulated-rigid-body}
\begin{algorithmic}
\STATE {\bfseries Input:} $\vq_{1:n}$, $\dot{\vq}_{1:n}$, $\vu_{1:n}$ 
\STATE {\bfseries Output:} $\ddot{q}_{0:n}$, $\bar{\va}_{0:n}$, $\bar{\vf}_{0:n}$
\FOR{$i=1$ {\bfseries to} $n$}
\STATE // Forward Kinematics\;
\STATE $\bar{\vv}_i = Ad_{T^{-1}_{\lambda, i}}\bar{\vv}_{\lambda} {+} \vs_i \dot{q}_i$ \;
\STATE $\bar{\bm{\eta}}_i = ad_{\bar{\vv}_i} \vs_i \: \dot{q}_i$
\ENDFOR
\FOR{$i=n$ {\bfseries to} $1$}
\STATE // Compute lumped inertia
\STATE$\bar{\mM}_{i:n} {=} \bar{\mM}_i {+} \sum_{k \in \mu} Ad_{T_{i,k}^{-1}}^* \bar{\Pi}_k Ad_{T_{i,k}^{-1}}$\;
\STATE// Compute bias forces \;
\STATE $\bar{\vf}_{b, i} = - ad_{\bar{\vv}_i}^* \bar{\mM}_{i:n} \bar{\vv}_i {+} \sum_{k \in \mu} Ad_{T_{i,k}^{-1}}^* \left( \bar{\vf}_{b, k} {+} \bar{\bm{\beta}}_k \right)$
\STATE $\bar{\Psi}_i = \left(\vs_i^T \: \bar{\mM}_{i:n} \: \vs_i\right)^{-1}$
\STATE $\bar{\bm{\Pi}}_i = \bar{\mM}_{i:n} {-} \bar{\mM}_{i:n} \vs_i \bar{\Psi}_i \vs_i^T \bar{\mM}_{i:n}$
\STATE $\bar{\bm{\beta}}_i = \bar{\mM}_{i:n}
\left(\bar{\bm{\eta}}_i{+}\vs_i \bar{\Psi}_i \left( u_i {-} \vs_i^T \left( \bar{\mM}_{i:n} \bar{\bm{\eta}}_i {+} \bar{\vf}_{b, i} \right) \right) \right)$
\ENDFOR
\FOR{$i=1$ {\bfseries to} $n$}
\STATE // Newton Euler Equations \;
\STATE $\ddot{q}_i = \bar{\Psi}_i \left( u_i {-} \vs_i^T \left(\bar{\mM}_{i:n} \left( Ad_{T_{\lambda, i}^{-1}} \bar{\va}_{\lambda} {+} \bar{\eta}_i \right) {-} \bar{\vf}_{b, i} \right)\right)$\;
\STATE $\bar{\va}_i = Ad_{T_{\lambda, i}^{-1}} \bar{\va}_{\lambda} {+} \vs_i \ddot{q}_i {+} \bar{\bm{\eta}}_i$\;
\STATE $\bar{\vf}_i = \bar{\mM}_{i:n} \bar{\va}_i {+} \bar{\vf}_{b, i}$\;
\ENDFOR
\end{algorithmic}
\end{algorithm}

\section*{Virtual Physical Parameters}
\noindent To enable learning via standard gradient descent methods, we carefully parameterize the physical parameters of the algorithm by a set of unrestricted, \emph{virtual} parameters \cite{ting2006bayesian, sutanto2020encoding} to result in physically correct model parameters. These virtual are necessary as not all link parameters are physically plausible, e.g., the link mass has to be positive. 

\subsection*{Kinematics}\label{sec:kin} \noindent
The transformation $\mT(q)$ between two links depends on link dimension, the joint position and the joint constraint connecting the two links. We decompose this transformation $\mT(q){=}\mT_{O} \mT_q(q)$ into a fixed transform $\mT_O$ and variable transform $\mT_q(q_i)$. The fixed transform describes the the distance and rotation between two joints and is parametrized by the translation vector $\vp_k$ and the RPY Euler angles $\vtheta_R{=}[\phi_x, \phi_y, \phi_z]^T$. The transformation is then described by 
\begin{align}
    \mT_O &= 
    \begin{bmatrix}
        \mR_z(\phi_z) \mR_y(\phi_y) \mR_x(\phi_x)  & \vp_k \\
        0   & 1
    \end{bmatrix}
\end{align}
where $\mR_a(\phi)$ denotes the rotation matrix corresponding to the rotation by $\phi$ about axis $a$ using the right-hand rule. Note that the rotation matrices about the elementary axis only depend on $\vtheta_R$ through arguments to trigonometric functions. Due to the periodic nature of those functions we obtain a desired unrestricted parameterization. The variable transform $\mT_q(q_i)$ describes the joint constraint and joint configuration. For all joints we assume that the variable link axis is aligned with the z-axis. Hence, the transformation matrix and joint motion vector for revolute joints ($\mT_{q_{r}},\vs_r$) and prismatic joints ($\mT_{q_{p}},\vs_p$) is described by
\begin{alignat*}{2}
    &\mT_{q_r} = 
    \begin{bmatrix}
        \mR_{z}(q) & \mathbf{0} \\
        \mathbf{0}   & 1
    \end{bmatrix}, \hspace{25pt}
    \mT_{q_p} &&= 
    \begin{bmatrix}
        \mathbf{0} &  q \: \ve_z\\
        \mathbf{0}   & 1
    \end{bmatrix}, \\
    &\:\:\vs_r = \: \begin{bmatrix}
        \mathbf{0} &  \ve_z
    \end{bmatrix}, \hspace{48pt}
    \vs_p &&= \begin{bmatrix}
        \ve_z & \mathbf{0}
    \end{bmatrix}, 
\end{alignat*}
with the $z$ axis unit vector $\ve_z$.
Technically, one can also use fixed joints with $\mT_q{=}\mathbf{I}$ but this simply yields an overparameterized model and is not necessary for describing the system dynamics. The complete kinematics parameters of a link are summarized as $\vtheta_K{=}\{\vtheta_R, \vp_k \}$

\subsection*{Inertias} \noindent
For physical correctness, the diagonal rotational inertia $\mJ_p{=}\text{diag}([J_x, J_y, J_z])$ at the body's CoM and around principal axes must be positive definite and need to conform with the triangle inequalities~\cite{traversaro2016identification}, i.e.,
\begin{equation*}
J_x \leq J_y + J_z , \quad 
J_y \leq J_x + J_z , \quad
J_z \leq J_x + J_y \;.
\end{equation*}
To allow an unbounded parameterization of the inertia matrix, we introduce the parameter vector
$\vtheta_L{=}[\theta_{\sqrt{L_1}},\theta_{\sqrt{L_2}},\theta_{\sqrt{L_3}}]\tran$, 
where $L_i$ represents the central second moments of mass of the density describing the mass distribution of the rigid body with respect to a principal axis frame. Then rotational inertia is described by
\begin{align*}
    \mJ_p = \text{diag}(
\theta_{\sqrt{L_2}}^2{+}\theta_{\sqrt{L_3}}^2, \:
\theta_{\sqrt{L_1}}^2{+}\theta_{\sqrt{L_3}}^2, \:
\theta_{\sqrt{L_1}}^2{+}\theta_{\sqrt{L_2}}^2
).
\end{align*}
The rotational inertia is then mapped to the link coordinate frame using the parallel axis theorem described by 
\begin{equation}
    \mJ = \mR_J \mJ_p \mR_J\tran + m[\vp_m][\vp_m]
\end{equation}
with the link mass $m$ and the translation $\vp_m$ from the coordinate from to the CoM. 
The fixed affine transformation uses the same parameterization as described in \ref{sec:kin}. 
The mass of the rigid body is parameterized by $\theta_{\sqrt{m}}$ where $m{=}\theta_{\sqrt{m}}^2$.
Given the dynamics parameters $\vtheta_{I}{=}\{\vtheta_L, \theta_{\sqrt{m}}, \vtheta_J, \vp_m\}$ for each link, the inertia in the desired frame using as well as generalized inertia $\bar{\mM}$ can be computed.


\end{document}